\documentclass[10pt,twocolumn,letterpaper]{article}

\usepackage[pagenumbers]{cvpr} 

\newcommand{\TODO}[1]{\textbf{\color{red}[TODO: #1]}}
\renewcommand{\TODO}[1]{}

\definecolor{cvprblue}{rgb}{0.21,0.49,0.74}
\usepackage[pagebackref,breaklinks,colorlinks,allcolors=cvprblue]{hyperref}

\def\paperID{*****} 
\def\confName{CVPR}
\def\confYear{2026}

\title{Mine-JEPA: In-Domain Self-Supervised Learning for Mine-Like Object Classification in Side-Scan Sonar}

\author{
Taeyoun Kwon$^{1,2}$ \quad
Youngwon Choi$^{1}$ \quad
Hyeonyu Kim$^{1}$ \quad
Myeongkyun Cho$^{1,3}$ \\
Junhyeok Choi$^{1}$ \quad
Moon Hwan Kim$^{1,4}$ \\[0.3em]
{\normalsize
$^{1}$Maum AI Inc. \quad
$^{2}$Seoul National University \quad
$^{3}$KAIST \quad
$^{4}$Yonsei University} \\[0.2em]
{\tt\small \{taeyoun.kwon, youngwonchoi, mkcho, acensia, hykim, mhk\}@maum.ai}
}

\begin{document}
\maketitle
\begin{abstract}
Side-scan sonar (SSS) mine classification is a challenging maritime vision problem characterized by extreme data scarcity and a large domain gap from natural images. While self-supervised learning (SSL) and general-purpose vision foundation models have shown strong performance in general vision and several specialized domains, their use in SSS remains largely unexplored. We present Mine-JEPA, the first in-domain SSL pipeline for SSS mine classification, using SIGReg, a regularization-based SSL loss, to pretrain on only 1,170 unlabeled sonar images. In the binary mine vs.\ non-mine setting, Mine-JEPA achieves an F1 score of 0.935, outperforming fine-tuned DINOv3 (0.922), a foundation model pretrained on 1.7B images. For 3-class mine-like object classification, Mine-JEPA reaches 0.820 with synthetic data augmentation, again outperforming fine-tuned DINOv3 (0.810). We further observe that applying in-domain SSL to foundation models degrades performance by 10--13 percentage points, suggesting that stronger pretrained models do not always benefit from additional domain adaptation. In addition, Mine-JEPA with a compact ViT-Tiny backbone achieves competitive performance while using 4$\times$ fewer parameters than DINOv3. These results suggest that carefully designed in-domain self-supervised learning is a viable alternative to much larger foundation models in data-scarce maritime sonar imagery.
\end{abstract}    
\section{Introduction}
\label{sec:intro}

Underwater mines pose a severe threat to ships and submarines, making effective mine countermeasures (MCM) essential for naval operations, which typically follow a pipeline of detection, classification, identification, and disposal~\cite{hozyn2021review}.
In this process, side-scan sonar (SSS)~\cite{blondel2010handbook} plays a central role in the first two stages by surveying wide seabed areas from platforms such as ships or autonomous underwater vehicles (AUVs). Traditionally, SSS imagery has been analyzed manually by trained operators, making the process labor-intensive and time-consuming, and thereby motivating the development of automated methods for mine detection and classification~\cite{steiniger2022survey}. While machine learning and deep learning methods have increasingly been applied to this task, most existing approaches remain heavily dependent on supervised learning due to the scarcity of labeled sonar data. As a result, representation-learning approaches that have driven progress in natural-image vision, including self-supervised learning (SSL) and foundation models, have received little attention in SSS imagery.

\begin{figure}[t]
    \centering
    \includegraphics[width=0.93\columnwidth]{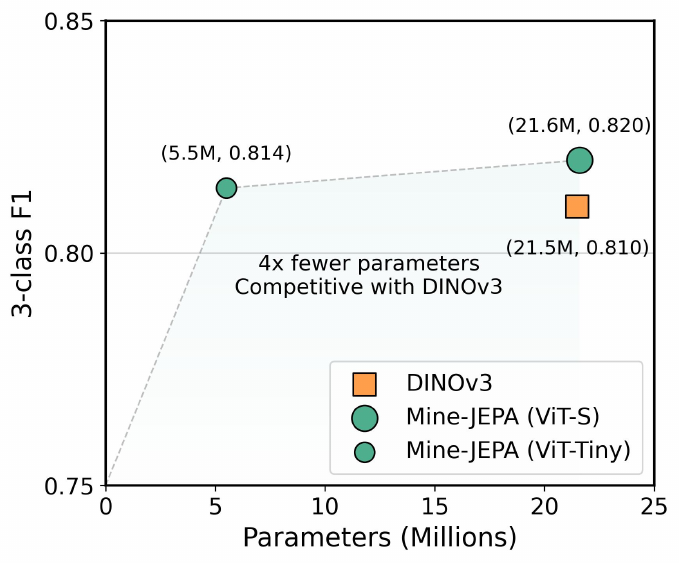}
    \caption{\textbf{Performance comparison across model scales.} Using only 1,170 sonar images for pretraining, Mine-JEPA outperforms DINOv3, pretrained on 1.7B images, at both backbone scales. ViT-Tiny remains competitive with $4\times$ fewer parameters.}
    \label{fig:intro}
\end{figure}

A natural approach to bridge this gap is in-domain SSL. While such strategies are increasingly adopted in other imaging domains, such as medical imaging and remote sensing (Sec.~\ref{sec:related_work_23}), they remain largely absent in SSS mine classification. We identify two key challenges underlying this gap. First, there is a substantial domain mismatch between natural-image pretraining and SSS imagery: vision models are typically trained on large-scale natural image datasets, whereas SSS interpretation relies on acoustic highlight–-shadow structures and seabed texture statistics that differ fundamentally from natural images. Second, the limited size of SSS datasets undermines the assumptions behind many successful SSL strategies, which are primarily developed and validated in large-scale settings, making representation learning particularly challenging in this extreme small-data regime.

In this work, we focus on the mine classification stage of the SSS analysis pipeline and investigate in-domain self-supervised pretraining for this task. Our approach builds on LeJEPA~\cite{balestriero2025lejepa}, a recent SSL framework that employs the Sketched Isotropic Gaussian Regularization (SIGReg) objective to enable stable representation learning in limited-data regimes without requiring teacher--student architectures. We adapt this framework to a public side-scan sonar dataset~\cite{santos2024side}, which contains 1,170 real sonar images, using domain-specific augmentations and a lightweight projection design, while initializing the backbone from ImageNet-1K pretrained weights.  Empirically, we show that combining SIGReg with SSS-adapted augmentation and ImageNet-1K initialization enables a lightweight SSL pipeline to match or outperform large-scale vision foundation models pretrained on natural-image datasets. We further observe that continuing SSL on domain-specific data from strong foundation models can degrade representation quality, highlighting the importance of appropriate initialization. Overall, our results suggest that in-domain SSL is a practical pretraining strategy for SSS mine classification, and highlight the effectiveness of LeJEPA/SIGReg in extreme small-data regimes.
\section{Related Work}
\label{sec:related_work}

\subsection{Side-Scan Sonar Mine Classification}
Early studies on side-scan sonar mine classification primarily relied on hand-crafted features extracted from regions of interest (ROIs), followed by conventional classifiers such as support vector machines (SVMs)~\cite{hollesen2011comparison} and boosting-based methods~\cite{sawas2010cascade}. Representative features included highlight--shadow geometry, template similarity, Haar-like features~\cite{sawas2010cascade}, local binary patterns (LBP)~\cite{barngrover2014semisynthetic}, and SIFT-style local descriptors~\cite{zhu2014model}. While these approaches established an initial automated classification pipeline, they remained dependent on manual feature design and often showed limited robustness in data-scarce settings~\cite{hozyn2021review}.

More recent work has shifted toward deep learning-based classification, especially convolutional neural networks (CNNs)~\cite{Gebhardt2017HuntingFN, dzieciuch2016non, 8600472}, which replace manual feature engineering with learned representations from sonar image patches or ROIs. Subsequent work increasingly adopted transfer learning from natural-image models~\cite{mckay2017s, 8945982}, using architectures such as AlexNet and VGG as feature extractors or fine-tuning them for mine and mine-like object recognition. However, prior work has remained largely within supervised or transfer-learning settings, and SSL-pretrained backbones have not been systematically investigated for SSS mine classification.

\subsection{Self-Supervised Learning for Vision}
Self-supervised learning (SSL) for vision has developed along several major paradigms. Contrastive methods such as SimCLR~\cite{chen2020simple} and MoCo~\cite{he2020momentum} learn view-invariant representations by matching augmented views of the same image. Masked-image modeling instead learns from reconstructing missing patches, as exemplified by MAE~\cite{he2022masked}. Self-distillation and predictive approaches, including BYOL~\cite{grill2020byol}, DINO~\cite{caron2021dino}, and I-JEPA~\cite{assran2023ijepa}, further show that strong visual features can emerge without explicit negative pairs.

A complementary line of work addresses representation collapse through explicit regularization. Barlow Twins~\cite{zbontar2021barlow} reduces redundancy through cross-correlation alignment, whereas VICReg~\cite{bardes2021vicreg} combines invariance with variance and covariance constraints. More recently, LeJEPA introduces SIGReg in a lean JEPA formulation that avoids teacher--student architectures and is particularly appealing in limited-data settings~\cite{balestriero2025lejepa}. This design is especially relevant to our setting, where dataset size is limited and stable optimization is a central concern. At the large-scale end, DINOv2 and DINOv3 demonstrate the strength of SSL when trained on massive curated datasets~\cite{oquab2024dinov2,simeoni2025dinov3}. Prior work suggests that adapting strong pretrained representations under substantial distribution shift may degrade transfer performance~\cite{kumarfine}. This concern is especially relevant when transferring natural-image models to sonar imagery, where both the image statistics and the data regime differ from those of typical large-scale pretraining.

\subsection{Domain-Specific SSL \& the Remaining Gap}
\label{sec:related_work_23}
Domain-specific SSL has been explored in several neighboring visual domains, including 3D medical imaging~\cite{zhou2021models}, remote sensing~\cite{9710545}, and marine image analysis~\cite{ciranni2025domain}. These studies suggest that in-domain pretraining can be beneficial when labeled data are limited, particularly when the learning objective and augmentation strategy are matched to domain-specific image statistics. Motivated by these advances, we investigate whether similar gains can be achieved for SSS mine classification. In this setting, domain-specific SSL remains largely unexplored despite the domain gap from natural-image pretraining and the limited size of available datasets. Our work addresses this gap by evaluating in-domain SSL for sonar imagery and comparing it against both standard initialization baselines and large-scale pretrained models.

\begin{figure*}[t!]
    \centering
    \includegraphics[width=\textwidth]{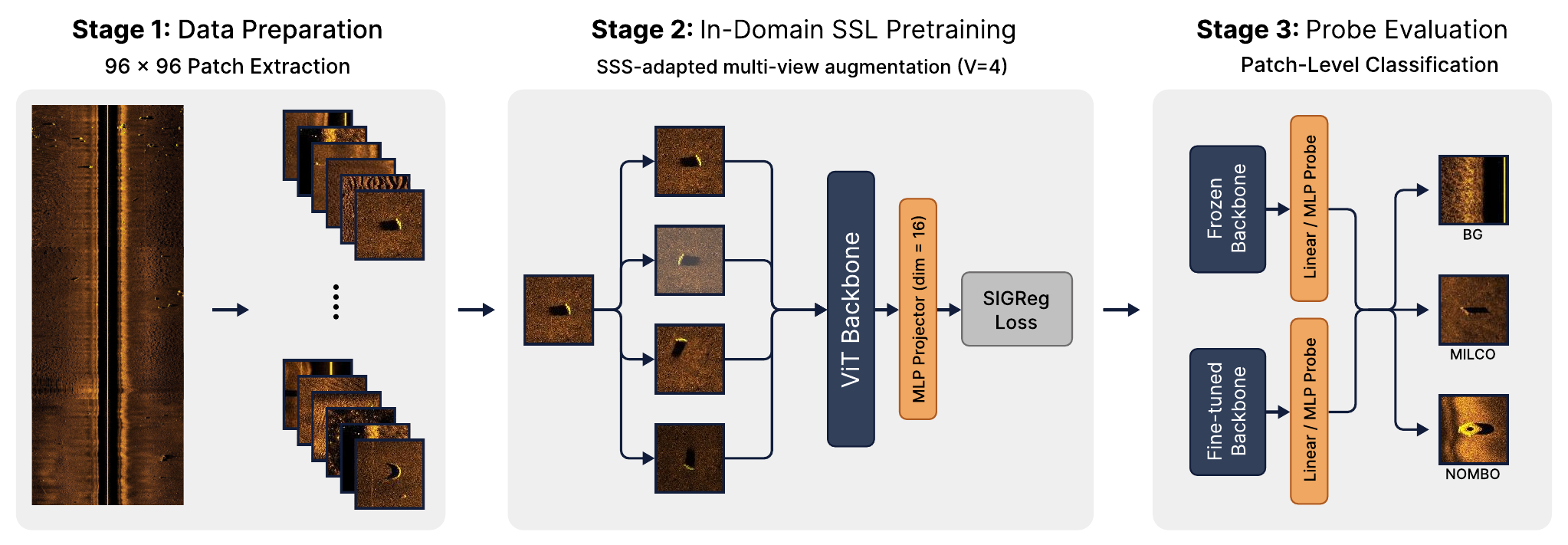}
    \caption{\textbf{Overview of the Mine-JEPA pipeline.} Stage 1 extracts unlabeled $96\times96$ patches from full SSS images using a sliding window. Stage 2 performs in-domain self-supervised pretraining with SSS-adapted multi-view augmentation and the SIGReg objective. Stage 3 evaluates the pretrained backbone with frozen or fine-tuned probes for downstream patch classification.}
    \label{fig:method}
\end{figure*}

\section{Method}
Mine-JEPA is designed for settings in which labeled SSS data are extremely scarce, while unlabeled sonar imagery can still provide meaningful learning signal. To address this challenge, we adopt a three-stage pipeline consisting of data preparation, in-domain SSL pretraining on unlabeled SSS patches, and probe evaluation on labeled patches. In this section, we describe the problem formulation and overall pipeline (Sec.~\ref{sec:method_pipeline}), introduce the SIGReg loss that enables stable SSL in a setting with only 1,170 source images (Sec.~\ref{sec:sigreg}), and present the SSS-specific adaptations that bridge the gap between natural-image SSL and sonar imagery (Sec.~\ref{sec:sss_adapted}).

\subsection{Problem Formulation \& Pipeline}
\label{sec:method_pipeline}
Our goal is to classify $96\times96$ patches extracted from side-scan sonar (SSS) images into three categories defined by the dataset taxonomy: background (BG), mine-like contact (MILCO), and non-mine bottom object (NOMBO). We also evaluate a binary formulation (mine vs.\ non-mine) by merging BG and NOMBO.

Mine-JEPA follows a three-stage pipeline (Fig.~\ref{fig:method}). In Stage 1 (data preparation), we extract approximately 153K unlabeled $96\times96$ patches from 1,170 SSS images with stride 64. In Stage 2 (in-domain SSL pretraining), we pretrain a Vision Transformer (ViT)~\cite{dosovitskiy2021image} backbone with the SIGReg loss (Sec.~\ref{sec:sigreg}), together with the SSS-adapted augmentation and initialization strategies described in Sec.~\ref{sec:sss_adapted}. In Stage 3 (probe evaluation), we attach a lightweight linear or MLP head to the pretrained backbone and evaluate it under either frozen or fine-tuned settings. The evaluation protocol and dataset details are described in Sec.~\ref{sec:data_setup}.

\subsection{SIGReg Self-Supervised Learning}
\label{sec:sigreg}
A central challenge in this setting is to obtain stable SSL under severe data scarcity. Many high-performance SSL methods rely on teacher--student architectures, momentum encoders, large batch sizes, or carefully tuned collapse-prevention heuristics. Although such designs can be effective at large scale, they may become unstable or unnecessarily complex when only 1,170 source images are available~\cite{elnouby2021large}. We therefore build Mine-JEPA on SIGReg, a regularization-based SSL objective introduced in LeJEPA~\cite{balestriero2025lejepa}, as a simpler and more stable alternative for small-data representation learning.

Given a batch of patches, we generate multiple augmented views from each patch and encode them using a backbone and an MLP projector. Let \(z_{i,v}\) denote the projected embedding of the view \(v\) for the patch \(i\), and let \(\bar{z}_i\) denote the mean embedding across views of the same patch. The training objective combines an invariance term and a regularization term:

\[
\mathcal{L} = (1-\lambda)\mathcal{L}_{\text{inv}} + \lambda \mathcal{L}_{\text{sig}}
\]

The invariance loss encourages different augmented views of the same patch to converge to similar representations:

\[
\mathcal{L}_{\text{inv}} = \frac{1}{NV}\sum_{i=1}^{N}\sum_{v=1}^{V}\|z_{i,v} - \bar{z}_i\|_2^2
\]

The SIGReg loss regularizes the overall embedding distribution toward a standard normal distribution by computing an Epps-Pulley-style goodness-of-fit statistic~\cite{epps1983test} through random slicing. This regularization prevents representation collapse without requiring negative pairs, exponential moving average (EMA) teachers, or centering and sharpening heuristics. Such simplicity is particularly attractive in the small-data regime, where ease of optimization and low hyperparameter burden are important. Moreover, the loss is computed with \(O(N)\) complexity, making it practical under limited-data and small-batch settings.

An important design choice in Mine-JEPA is the use of a very low projection dimension (\(d=16\)). We treat this not as a mere implementation detail, but as an intentional structural bottleneck. Under limited data, a low-dimensional projection space can regularize representation learning and reduce the tendency of overly large projection heads to overfit. As shown later in Sec.~\ref{sec:comparison}, competitive performance remains achievable even under this highly compressed setting. Therefore, the combination of invariance and distributional regularization provides a simple and stable SSL objective well suited to scarce SSS data.

\subsection{Mine-JEPA: SSS-Adapted SSL}
\label{sec:sss_adapted}
While SIGReg provides the core SSL objective, effective pretraining in SSS also requires the overall learning pipeline to be matched to the characteristics of sonar imagery. Standard SSL recipes are designed for natural RGB images and therefore do not necessarily transfer well to acoustic imagery. We accordingly adapt Mine-JEPA to the target domain along three complementary dimensions: augmentation, initialization, and data composition.

\textbf{SSS-Adapted Augmentation.} Standard SSL augmentation pipelines typically include strong color jitter, hue and saturation variation, random solarization, and random grayscale conversion, which can be effective for natural RGB imagery. However, such transformations can be ineffective or even harmful for SSS, where chromatic information is minimal and acoustic backscatter intensity is represented in false color. We therefore restrict color jitter to brightness and contrast only (saturation=0, hue=0), and remove solarization and grayscale conversion. In contrast, we add vertical flip and random rotation (±15°) to reflect directional invariances in sonar scanning geometry, while retaining horizontal flip, random resized crop (scale 0.5–1.0), and Gaussian blur. All images are normalized using dataset-specific statistics. This augmentation strategy is not merely an implementation detail, but a core part of the method: our experiments show that naive natural-image augmentation can substantially degrade SSS performance, whereas SSS-adapted augmentation significantly improves representation quality.

\textbf{Initialization Strategy.} Initialization is particularly important in the small-data regime. Rather than training entirely from scratch, we initialize the ViT backbone with ImageNet-1K (IN1K) pretrained weights and then apply domain SSL on top, while always keeping the projector head randomly initialized. This strategy allows the model to start from general visual features such as edges and textures, while still enabling SIGReg to adapt them to the SSS domain. It also reflects a practical middle ground. ImageNet initialization provides useful low-level visual priors, while avoiding the highly specialized representation structure of large foundation models already optimized for broad natural-image transfer. As shown later in Sec.~\ref{sec:init_strategy}, IN1K provides a more suitable basis for domain adaptation than either random initialization or DINOv3 initialization, while the stronger pretrained model DINOv3 instead undergoes representation degradation under domain SSL.

\textbf{Data Composition.} Finally, we investigate whether additional unlabeled data that are not identical to the target dataset, but remain structurally similar to sonar imagery, can benefit in-domain SSL. In addition to the approximately 153K real patches extracted from the public sonar dataset, we consider a Real+Syn setting that adds approximately 256K synthetic sonar patches, resulting in roughly 409K total patches. This synthetic-real mixture is intentionally heterogeneous: the real patches are RGB, whereas the synthetic sonar patches are grayscale. Rather than treating this mismatch as noise to avoid, we use it as an opportunity to test whether SSL can absorb sonar-like structure even in the presence of a moderate modality gap. This question is practically important because synthetic or auxiliary unlabeled sonar-like data may be much easier to scale than collecting new labeled mine data.

Each subset is normalized using its own statistics and then concatenated for pretraining. As shown in our experiments, this heterogeneous data composition leads to consistent performance gains. This suggests that combining SSS-adapted augmentation with heterogeneous sonar-like data can improve representation quality even when the source distribution is not perfectly matched.

\begin{figure}[t!]
    \centering
    \includegraphics[width=\columnwidth]{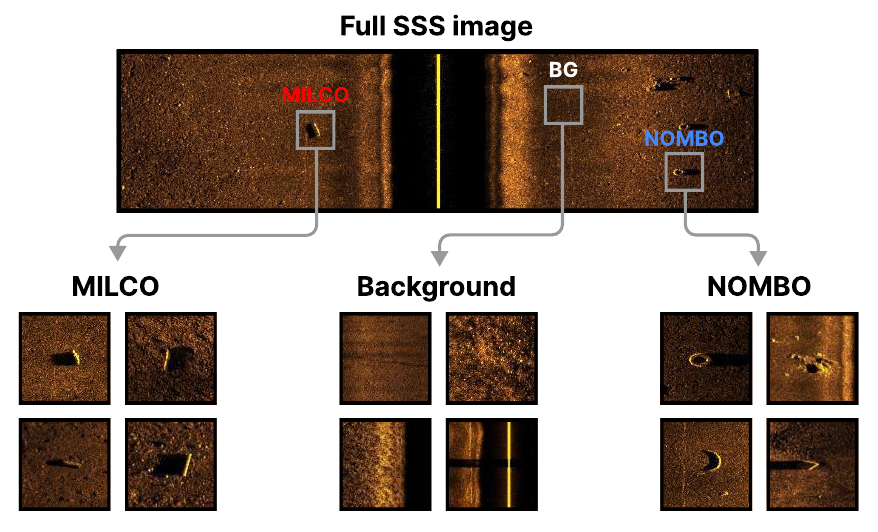}
    \caption{\textbf{Representative samples from the public side-scan sonar dataset.} Top: a full SSS image with annotated mine-like contacts (MILCO) and non-mine bottom objects (NOMBO). Bottom: $96\times96$ patches from the three classes used in our evaluation: MILCO, background (BG), and NOMBO.}
    \label{fig:dataset_samples}
\end{figure}

\begin{table*}[t!]
\centering
\small
\renewcommand{\arraystretch}{1.08}
\setlength{\tabcolsep}{7pt}
\caption{\textbf{3-class patch classification on BG/MILCO/NOMBO.} For each method, we report the best probe mode and the corresponding macro-F1, NOMBO F1, and accuracy. Probe modes: \textit{linear} and \textit{mlp} (frozen backbone), \textit{finetune} and \textit{ft mlp} (fine-tuned backbone).}
\label{tab:three_class_patch}
\begin{tabular*}{0.96\textwidth}{@{\extracolsep{\fill}}lcccccc}
\toprule
\textbf{Method} & \textbf{Init} & \textbf{SSL Data} & \textbf{Best Probe} & \textbf{macro-F1} & \textbf{NOMBO F1} & \textbf{Acc} \\
\midrule
Random Init & Random & —        & ft\_mlp  & 0.557$\pm$0.052 & 0.439$\pm$0.091 & 58.3$\pm$4.0\% \\
IN1K Only   & IN1K   & —        & ft\_mlp  & 0.739$\pm$0.028 & 0.647$\pm$0.046 & 76.3$\pm$2.6\% \\
DINOv3      & DINOv3 & —        & ft\_mlp  & 0.810$\pm$0.025 & 0.700$\pm$0.068 & 83.4$\pm$2.0\% \\
SimCLR      & IN1K   & Real+Syn & finetune & 0.801$\pm$0.038 & 0.693$\pm$0.084 & 82.6$\pm$2.6\% \\
VICReg      & IN1K   & Real+Syn & finetune & 0.800$\pm$0.037 & 0.676$\pm$0.068 & 82.8$\pm$3.0\% \\
BYOL        & IN1K   & Real+Syn & finetune  & 0.693$\pm$0.028 & 0.539$\pm$0.059 & 72.8$\pm$2.4\% \\
Mine-JEPA   & IN1K   & Real+Syn & finetune & \textbf{0.820$\pm$0.018} & \textbf{0.734$\pm$0.026} & \textbf{83.8$\pm$1.8\%} \\
\bottomrule
\end{tabular*}
\end{table*}

\section{Experiments}
We evaluate Mine-JEPA on a publicly available side-scan sonar benchmark introduced by Santos et al.~\cite{santos2024side}. After describing the experimental setup (Sec.~\ref{sec:data_setup}), we present results for both 3-class and binary classification (Sec.~\ref{sec:main_results}). We then analyze key design choices, including initialization strategy and SSL method selection (Sec.~\ref{sec:analysis}).

\subsection{Dataset \& Setup}
\label{sec:data_setup}
\textbf{Dataset and SSL Pretraining Data.} We use a publicly available side-scan sonar benchmark for mine detection introduced by Santos et al.~\cite{santos2024side}. It contains 1,170 images, of which 304 contain 668 annotated objects (437 MILCO and 231 NOMBO), and we use an 80/10/10 train/val/test split stratified by annotation presence. For self-supervised pretraining, we extract $96\times96$ patches from all 1,170 images with stride 64, yielding approximately 153K unlabeled patches. In the Real+Syn setting, we additionally include approximately 256K synthetic sonar patches, resulting in approximately 409K total patches. Real and synthetic patches are normalized using their own statistics. This setup allows us to evaluate learning from scarce real sonar imagery and the effect of adding sonar-like synthetic data.

\textbf{Probe Evaluation Protocol.} We evaluate learned representations on labeled patches extracted from annotated regions. For 3-class classification (BG/MILCO/NOMBO), the train/test splits contain 891/110 patches; for binary mine vs.\ non-mine classification, they contain 706/86 patches. We consider four probe modes: linear, mlp, finetune, and finetune\_mlp. In the frozen setting, only the classification head is trained, whereas in the fine-tune setting the backbone is updated jointly with the head. All probes are trained with AdamW, cosine annealing, and early stopping, and all results are reported as mean$\pm$std over 10 random seeds using macro-averaged F1 as the primary metric. This protocol allows us to compare representation quality under both frozen-feature transfer and full end-to-end adaptation.

\begin{table}[t!]
\centering
\small
\renewcommand{\arraystretch}{1.1}
\setlength{\tabcolsep}{7pt}
\caption{\textbf{Default SSL pretraining configuration (ViT-S/16).} All architectures use the same setup unless noted otherwise.}
\label{tab:ssl_config}
\begin{tabular}{lc}
\toprule
\textbf{Hyperparameter} & \textbf{Value} \\
\midrule
Projection dim. & 16 \\
$\lambda$       & 0.1 \\
Views ($V$)     & 4 \\
Batch size      & 1024 \\
Optimizer       & AdamW (weight decay 0.05) \\
Learning rate   & $1.4\times10^{-3}$ \\
Schedule        & 1-epoch warmup + cosine decay \\
Epochs          & 100 \\
\bottomrule
\end{tabular}
\end{table}

\begin{table*}[t!]
\centering
\small
\renewcommand{\arraystretch}{1.08}
\setlength{\tabcolsep}{8pt}
\caption{\textbf{Model comparison across 3-class and binary settings.} Mine-JEPA uses Real+Syn data for 3-class classification and Real data for binary classification. For each method, we report the best probe mode.}                                                 
\label{tab:model_comparison}
\begin{tabular*}{0.92\textwidth}{@{\extracolsep{\fill}}lcccccc}
\toprule
\textbf{Method} & \textbf{Init} & \textbf{Params} & \textbf{3-class F1} & \textbf{2-class F1} & \textbf{MILCO Recall (\%)} \\
\midrule
DINOv3               & DINOv3 & 21.5M & 0.810$\pm$0.025 & 0.922$\pm$0.018 & 88.1$\pm$1.3 \\
Mine-JEPA (ViT-S)    & IN1K   & 21.6M & \textbf{0.820$\pm$0.018} & \textbf{0.935$\pm$0.020} & 90.9$\pm$2.4 \\
Mine-JEPA (ViT-Tiny) & IN1K   & 5.5M  & 0.814$\pm$0.026 & \textbf{0.935$\pm$0.020} & \textbf{91.4$\pm$2.3} \\
\bottomrule
\end{tabular*}
\end{table*}

\begin{table*}[t]
\centering
\small
\caption{\textbf{Ablation on initialization, augmentation, and data composition.} Real$^\ast$ denotes natural-image augmentation (saturation/hue jitter, solarization, grayscale). Rows 1--2 compare against Random Init, rows 3--4 against DINOv3, and rows 5--8 show cumulative gains; each $\Delta$ is relative to the preceding row.}
\label{tab:ablation}

\begin{tabular*}{0.92\textwidth}{@{\extracolsep{\fill}}lcccccc@{}}
\toprule
\textbf{Config} & \textbf{Init} & \textbf{SSL Data} & $\boldsymbol{\lambda}$ & \textbf{macro-F1} & $\boldsymbol{\Delta}$ \\
\midrule
Random Init         & Random & —         & —    & 0.557$\pm$0.052 & — \\
Natural Aug SSL     & Random & Real$^\ast$ & 0.02 & 0.312$\pm$0.040 & $-24.5$\%p \\
DINOv3 + SIGReg     & DINOv3 & Real      & 0.02 & 0.706$\pm$0.068 & $-10.4$\%p vs DINOv3 \\
DINOv3 + SIGReg     & DINOv3 & Real+Syn  & 0.02 & 0.677$\pm$0.031 & $-13.3$\%p vs DINOv3 \\
SIGReg (SSS aug) & Random & Real      & 0.02 & 0.725$\pm$0.021 & — \\
+ IN1K init         & IN1K   & Real      & 0.02 & 0.756$\pm$0.054 & +3.1\%p vs Random SSL \\
+ $\lambda$ tuning  & IN1K   & Real      & 0.1  & 0.799$\pm$0.036 & +4.3\%p \\
+ Add Real+Syn data & IN1K   & Real+Syn  & 0.1  & \textbf{0.820$\pm$0.018} & +2.1\%p \\
\bottomrule
\end{tabular*}
\end{table*}

\subsection{Main Results}
\label{sec:main_results}

We first present 3-class results, followed by comparisons on the binary setting and lightweight backbones.

Table~\ref{tab:three_class_patch} summarizes 3-class patch classification results. Random initialization establishes a lower bound (F1=0.557), IN1K pretraining improves performance to 0.739, and DINOv3 provides a strong baseline at 0.810. Under the same IN1K initialization, SimCLR (0.801) and VICReg (0.800) approach DINOv3, whereas BYOL degrades to 0.693. Mine-JEPA achieves the best overall performance in the Real+Syn setting, reaching 0.820. On the most difficult class, NOMBO, Mine-JEPA also improves over DINOv3 (0.734 vs. 0.700) with lower variance, indicating more stable recognition.

Table~\ref{tab:model_comparison} compares backbone choice across the 3-class and binary settings. In the 3-class setting, both ViT-S (0.820) and ViT-Tiny (0.814) match or outperform DINOv3 (0.810), while ViT-Tiny remains competitive despite using about 75\% fewer parameters (5.5M vs. 21.5M). In the binary setting, Mine-JEPA reaches F1=0.935 with higher MILCO recall (90.9--91.4\% vs. 88.1\%), suggesting the potential to reduce missed mines. The strong performance of ViT-Tiny in both settings further highlights the practical viability of lightweight backbones for resource-constrained platforms such as AUVs.

\subsection{Analysis}
\label{sec:analysis}
We analyze two factors that strongly influence SSL performance in the SSS domain. First, we examine initialization strategy and show that domain SSL can either improve or degrade performance depending on the starting representation. In particular, while IN1K initialization enables cumulative gains under in-domain SSL, applying the same adaptation procedure to a strong foundation model can instead harm performance (Sec.~\ref{sec:init_strategy}). Second, we compare alternative SSL objectives under matched settings and show that method choice substantially affects the ability to exploit heterogeneous synthetic data (Sec.~\ref{sec:comparison}).

\subsubsection{Initialization \& Adaptation Behavior}
\label{sec:init_strategy}
Table~\ref{tab:ablation} shows that the effect of domain SSL depends strongly on initialization. While the same adaptation procedure degrades DINOv3, it yields steady cumulative gains under IN1K initialization. The table also highlights that SSS-adapted augmentation is a necessary prerequisite for effective SSL in SSS.

\textbf{SSS-adapted augmentation as a prerequisite.} Applying natural-image augmentation—including saturation/hue jitter, solarization, and grayscale conversion—to SSS imagery yields F1=0.312, which is even worse than using no pretraining at all (0.557). In contrast, switching to the SSS-adapted augmentation described in Sec.~\ref{sec:sss_adapted} raises performance to 0.725. This large gap indicates that augmentation is not a minor implementation detail in SSS, but a central design choice that must respect the acoustic statistics and geometric invariances of sonar imagery.

\textbf{Foundation-model degradation under domain SSL.} When SIGReg is applied to DINOv3, performance drops from 0.810 to 0.706 on real data alone ($-10.4$\%p), and further drops to 0.677 ($-13.3$\%p) when synthetic data are added. This suggests that the representation learned by a large natural-image foundation model is not necessarily improved by additional domain SSL under strong distribution shift. Instead, adaptation can distort an already strong representation, consistent with prior observations on feature degradation under mismatched transfer settings~\cite{kumarfine}.

\textbf{Cumulative gains under IN1K initialization.} In contrast, starting from a less specialized IN1K initialization leads to steady gains under the same SSL recipe. Relative to the random-init SSL baseline (0.725), IN1K initialization improves performance to 0.756 ($+3.1$\%p), tuning $\lambda$ further raises it to 0.799 ($+4.3$\%p), and adding Real+Syn data yields the best overall result of 0.820 ($+2.1$\%p). These gains suggest that ImageNet-1K provides useful low-level visual priors while still leaving enough flexibility for effective adaptation to sonar imagery. In other words, in our setting, a moderately pretrained initialization is a better starting point for domain SSL than a stronger but more specialized foundation model.

Overall, Table~\ref{tab:ablation} suggests that successful adaptation in SSS depends not only on the SSL objective itself, but also on how augmentation, initialization, and data composition interact. Mine-JEPA performs best when these components are jointly matched to the specific constraints of the small-data sonar regime, highlighting the importance of coordinated design choices.

\begin{table*}[t!]
\centering
\small
\caption{\textbf{SSL method comparison and data scalability.} All methods use ViT-S/16 with IN1K initialization, 100 epochs. macro-F1 on 3-class classification. $\Delta$ = Real+Syn $-$ Real only.}
\label{tab:ssl_method}

\begin{tabular*}{0.8\textwidth}{@{\extracolsep{\fill}}lccccc@{}}
\toprule
\textbf{SSL Method} & \textbf{Loss Type} & \textbf{proj.\ dim} & \textbf{Real only} & \textbf{Real+Syn} & $\boldsymbol{\Delta}$ \\
\midrule
SIGReg (Ours) & Regularization & 16   & 0.799$\pm$0.036 & 0.820$\pm$0.018 & +2.1\%p \\
VICReg        & Regularization & 2048 & 0.774$\pm$0.025 & 0.800$\pm$0.037          & +2.6\%p \\
SimCLR        & Contrastive    & 128  & 0.806$\pm$0.013 & 0.801$\pm$0.038          & $-0.5$\%p \\
BYOL          & Distillation   & 256  & 0.774$\pm$0.031 & 0.693$\pm$0.028          & $-8.1$\%p \\
\bottomrule
\end{tabular*}
\end{table*}

\subsubsection{SSL Method Comparison \& Data Scalability}
\label{sec:comparison}
Table~\ref{tab:ssl_method} compares four SSL methods from different loss families under identical conditions (ViT-S/16, IN1K initialization, 100 epochs), with the Real-only and Real+Syn settings placed side by side to assess how each method responds to heterogeneous synthetic data.

A clear pattern emerges across SSL families. Regularization-based methods benefit from additional synthetic data: SIGReg improves from 0.799 to 0.820 (+2.1\%p), and VICReg improves from 0.774 to 0.800 (+2.6\%p). By contrast, the contrastive method SimCLR slightly declines from 0.806 to 0.801 ($-0.5$\%p), while the distillation-based method BYOL drops substantially from 0.774 to 0.693 ($-8.1$\%p). These results suggest that heterogeneous synthetic data are not uniformly beneficial, and that the robustness of the SSL objective to modality mismatch matters in this setting.

One possible explanation is that the gap between RGB real patches and grayscale synthetic patches interacts differently with different loss families. For SimCLR, such mismatch may weaken the usefulness of positive and negative comparisons, leading to only marginal gains or slight degradation. For BYOL, it may destabilize the consistency relation between online and target views. In contrast, regularization-based objectives such as SIGReg and VICReg appear better able to absorb additional sonar-like structure without relying on stricter pairwise or teacher--student alignment assumptions.

Among the compared methods, SIGReg achieves the strongest performance in the Real+Syn setting (0.820). Notably, it does so with a highly compressed projection space (projection dimension of 16), a single hyperparameter $\lambda$, and no momentum encoder. This result is practically important: in a small-data SSS regime, the most effective SSL objective is not the most architecturally complex one, but the one that most stably exploits limited and heterogeneous unlabeled data.

Taken together, these comparisons suggest that regularization-based SSL offers the most robust path for in-domain adaptation in SSS. The benefit of Real+Syn data in Mine-JEPA is therefore not simply a consequence of having more samples, but of pairing additional sonar-like data with an SSL objective that can use them effectively.
\section{Discussion \& Limitations}
\label{sec:discussion}

Our results suggest that, in side-scan sonar, effective pretraining depends less on model scale alone than on how well the learning strategy is matched to the target domain. In this setting, a simple in-domain SSL pipeline built on SIGReg, SSS-adapted augmentation, and ImageNet-1K initialization matched, and in some cases outperformed, a much larger general-purpose foundation model. The consistent gains from synthetic augmentation further indicate that additional sonar-like unlabeled data can remain useful even when heterogeneous. Moreover, the strong performance of ViT-Tiny highlights the practical viability of this approach for resource-constrained platforms such as AUVs, where onboard classification could reduce reliance on time-consuming post-mission analysis.

At the same time, our results show that naively applying in-domain SSL to stronger pretrained models can be counterproductive. In particular, DINOv3 consistently degraded after additional domain SSL. This observation is consistent with prior findings on feature distortion under distribution shift, and suggests that stronger pretrained representations may be less amenable to further adaptation when the target domain is both small and substantially mismatched from the original pretraining distribution. Understanding the conditions under which foundation models can or cannot benefit from additional domain adaptation remains an important open question, particularly as such models continue to grow in scale and capability.

This study is limited by the scale and scope of the available benchmark. All experiments are conducted on a single public dataset consisting of only 1,170 sonar images, and the resulting test sets are relatively small. We also do not evaluate across a broader range of mine types, sonar systems, or operational conditions. While the consistent gains from synthetic sonar patches are encouraging, their broader utility should be validated on larger and more diverse benchmarks. Future work could address these limitations by evaluating on multi-sensor benchmarks and operational sonar data collected under diverse environmental conditions. Despite these limitations, our results provide concrete empirical evidence that carefully designed in-domain SSL is a promising pretraining strategy for data-scarce maritime sonar imagery.

\section{Conclusion}

We presented Mine-JEPA, an in-domain self-supervised learning pipeline for side-scan sonar mine classification under severe data scarcity and substantial domain shift from natural images. Across both binary and 3-class settings, Mine-JEPA matched or outperformed DINOv3 on this benchmark, despite being pretrained on only 1,170 unlabeled sonar images. We also found that naively applying domain SSL to stronger pretrained models can degrade representation quality rather than improve it.

These findings suggest that effective pretraining for maritime sonar imagery does not necessarily require ever larger foundation models. Instead, combining SIGReg with domain-matched augmentation and appropriate initialization provides a practical and effective recipe for data-scarce maritime vision tasks. Future work includes extending this approach beyond patch-level classification to sliding-window detection and evaluating transfer across additional sonar modalities such as synthetic aperture sonar (SAS) and forward-looking sonar (FLS).

{
    \small
    \bibliographystyle{ieeenat_fullname}
    \bibliography{main}
}


\end{document}